\documentclass[11pt]{article}
\usepackage[T1]{fontenc}
\usepackage[utf8]{inputenc}
\usepackage{authblk}
\usepackage{acl2015}
\usepackage{times}
\usepackage{url}
\usepackage{latexsym}
\usepackage{graphicx}
\usepackage{amsfonts}
\usepackage{booktabs}
\usepackage{subcaption}
\usepackage{gb4e}
\usepackage{cgloss4e}
\noautomath

\title{Assessing the Stylistic Properties of Neurally Generated Text in Authorship Attribution\\}

\author[1]{\textbf{Enrique Manjavacas}}
\author[2]{\textbf{Jeroen De Gussem}}
\author[1]{\textbf{Walter Daelemans}}
\author[1]{\textbf{Mike Kestemont}}
\affil[1]{University of Antwerp\\CLiPS\\\tt\{firstname,lastname\}@uantwerpen.be}
\affil[2]{Ghent University\\Department of History\\\tt{jedgusse.degussem@ugent.be}}

\date{}

\begin{document}

\maketitle

\begin{abstract}
Recent applications of neural language models have led to an increased interest in the automatic generation of natural language. However impressive, the evaluation of neurally generated text has so far remained rather informal and anecdotal. Here, we present an attempt at the systematic assessment of one aspect of the quality of neurally generated text. We focus on a specific aspect of neural language generation: its ability to reproduce authorial writing styles. Using established models for authorship attribution, we empirically assess the stylistic qualities of neurally generated text. In comparison to conventional language models, neural models generate fuzzier text that is relatively harder to attribute correctly. Nevertheless, our results also suggest that neurally generated text offers more valuable perspectives for the augmentation of training data.
\end{abstract}

\section{Introduction}
In his landmark paper `Computing Machinery and Intelligence`, Turing
\shortcite{turing1950} quoted Jefferson's `The Mind of
Mechanical Man' \shortcite{jefferson1949}: `Not until a machine can write a sonnet or compose a concerto because of thoughts and emotions felt, and not by the chance fall of symbols, could we agree that machine equals brain'. Strikingly, these early pioneers of modern AI considered the conscious
creation of literature as a significant milestone on the long road towards general AI. In recent years, the automated generation of text, such as literature, has received a significant impetus from research in the field of neural language modeling. A variety of recent studies have demonstrated that neural language models can be used to synthesize new (literary) text, even at the character-level.

To a surprising extent, neurally generated text seems to make an authentic
impression on readers, due to its ability to mimic 
certain properties of the text on which it was trained, without it 
degrading into in a mere reproduction or patchwork of verbatim passages in
it. In one particularly visible blog post, Karpathy \shortcite{karpathy2015} demonstrated how a relatively simple 
character-level recurrent neural network, when trained on Shakespeare's oeuvre,
was able to generate new, artificial text which, certainly in the eyes
of non-experts, undeniably displayed some Shakespearean qualities. This blog has 
inspired a wide array of other applications -- ranging from cooking recipes \cite{recipe} to Bach's sonatas \cite{bach}.

Much of this work has so far been published in the online blogosphere and the assessment of the quality of neurally generated text has often remained fairly informal and anecdotal, apart from a number of more empirically oriented studies, for instance in the field of hiphop lyric generation \cite{potashEA2015,malmiEA2015}. In this paper, we report an attempt at a systematic assessment of the properties of neurally generated text in the context of style-based authorship attribution in stylometry \cite{stamatatos2009}. We address the following research questions: (1) \emph{To which extent is the text, neurally generated on the basis of a single author's oeuvre, still attributable to the original input author?} and (2) \emph{To which extent is the neural generation of text useful for training data augmentation in stylometry, e.g. for authors for whom little reference data is available?} 

Below, we first present the model architectures underlying our text generation, comparing a modern neural architecture to a more conventional ngram-based language model. Next, we describe the Latin data set which we will use (\emph{Patrologia Latina}) and discuss our experimental set-up (authorship attribution). We go on to present our attribution results; in the discussion section, we interpret and visualize these results. We conclude by pointing out viable future improvements.

\section{Character-Level Text Generation}\label{sec:cltg}
We approach the task of text generation with character-level Language Models (LM). In short, a LM is a probabilistic model of linguistic sequences that, at each step in a sequence, assigns a probability distribution over the vocabulary conditioned
on the prefix sequence. More formally, a LM is defined by Equation \ref{eq:lm}, 
\begin{equation}
LM(w_t) = P(w_t|w_{t-n}, w_{t-(n-1)}, ..., w_{t-1})
\label{eq:lm}
\end{equation}
where $n$ refers to the scope of the model ---i.e. the length of the prefix sequence 
taken into account to condition the output distribution at step $t$.  By extension, a LM defines a generative model of sentences where the probability of a sentence is
defined by the following equation:
\begin{equation}
P(w_1, w_2, ..., w_n) = \prod_i^n P(w_t|w_1, ..., w_{t-1})
\label{eq:lm_sent}
\end{equation}
\noindent Given its generative nature, a LM can easily be used for text generation. We start by sampling from the output distribution at step $t$ and, then, we recursively feed back the sampled symbol, together with any other previous output, to condition the generative distribution at
step $t+1$. Equation \ref{eq:lm_gen} shows formally the text generation process for a symbol at
step $t$ where $w_{t-1}^{\prime}$ is the generated symbol at step $t-1$ and $S$ refers
to any given sampling method. 
\begin{equation}
w_{t}^{\prime} = S[P(w_{t}|w_{t-n}^{\prime}, w_{t-(n-1)}^{\prime}, ..., w_{t-1}^{\prime})]
\label{eq:lm_gen}
\end{equation}
\noindent An obvious approach towards sampling is to select the symbol that maximizes the probability of 
the entire generated sequence (\emph{argmax} decoding). For a large vocabulary (e.g. in the
case of a word-level LM), the search quickly becomes impractical and is usually approximated
by means of beam search (including the extreme case of using a beamsize equal to 1, which corresponds
to picking the most probable symbol at each step).
However, when used for generation, the \emph{argmax} decoding approach tends to yield repetitive and dull
sentences, and eventually runs into dead-end loops. Therefore, we instead sample from the LM's output
distribution at each step.

The sampling approaches discussed so far attempt to strike a trade-off  between variability and 
correctness -- in the sense of departure from regularities observed in the training data.
Beam-search decoding will tend to generate sentences that are more formally correct (e.g. more similar to the sentences observed in the training corpus), while generating very similar and monotonous output
in the presence of similar histories. Conversely, multinomial sampling will make the output diverge 
more from the original training data, and therefore produce a more varied output, but with a tendency
towards more grammatically incorrect sentences. Focusing on multinomial sampling, the described 
trade-off can be operationalized in form of a parameter $\tau$, mostly referred to as 
``temperature", that is in charge of modifying the skewness of the parameters of the multinomial 
distribution to encourage more or less variability in exchange for potentially less or more
formally correct output.\footnote{
	Given the multinomial parameters $p = \{p_1, p_2, ..., p_k\}$ for a vocabulary size of $V$,
	the ``freezing'' transformation $p_i^{\tau} = p_i^{\frac{1}{\tau}} / \sum_j^V p_j^{\frac{1}{\tau}}$
	will flatten the original distribution for higher values of $\tau$, thereby ensuring more variability
	in the output. Conversely, lower values of  $\tau$ will skew it, thereby facilitating the outcome of
	the originally more probable symbol. For $\tau$ values approaching 0, we recover the simple argmax
	decoding procedure of picking the highest probability symbol at each step. 
}

A further aspect of our LM approach to text generation is topical variation. 
In order to ensure that during generation the LM explores the topical
distribution present in the training data, we implement the following procedure.
After having generated a fixed number of sentences $s$, a sentence from the LM's training data is sampled uniformly and used to seed the generation of the next $s$ sentences. Finally, we force the LM to generate fully terminated sentences by including end-of-sentence symbols (EOS) during training time and discarding any output sentence that reaches a maximum number of characters $m$ without having
generated the EOS symbol -- thus, we consider the generation of a single sentence
finished whenever the EOS symbol is produced and we only generate sentences
with a maximum number of characters $m$. This is motivated by the fact that 
very long sentences tend to degenerate into poor-quality text. Our generative system displays a total of 3 generation hyper-parameters:
$\tau$ (sampling temperature), $s$ (reset seed every $s$ sentences)
and $m$ (maximum $m$ characters per sentence).

\subsection{Ngram-based Language Model}
So far, we have kept the definition of the LM agnostic with respect to its concrete implementation.
In the current study we compare two widely-used LM architectures -- an ngram-based LM (NGLM) and a
Recurrent Neural Network-based LM (RNNLM). An NGLM is basically a conditional probability table for Equation \ref{eq:lm} that is estimated
on the basis of the count data for ngrams of a given length $n$.
Typically, NGLMs suffer from a data sparsity problem since for a large enough value of $n$
many possible conditioning prefixes will not be observed in the training data and the
corresponding probability distribution will be missing.
To alleviate the sparsity problem, two techniques---smoothing and backoff models---can be
used that either reserve some probability mass and evenly redistribute it 
across unobserved
ngrams (smoothing) or resort back to a lower-order model to provide an approximation to
the conditional distribution of an unobserved ngram (backoff models).
Here, however, we implement an unsmoothed LM since we only use the LM for generation,
where it is not necessary to compute probabilities for unseen ngrams. 
An unsmoothed NGLM only has one model hyper-parameter---the ngram order.

\subsection{RNN-based Language Model}\label{sec:rnn}
A RNNLM implements a language model using a Recurrent Neural Network (RNN) to
allow left-to-right information flow during sequence processing \cite{Mikolov2010}.
As shown in \cite{Bengio2003},
at a given step $t$, a RNNLM \cite{Elman1990} (i) first computes a distributed
representation $w_t$ with dimensionality $M$ of the input $x_t$,
(ii) it then feeds the resulting vector into an RNN layer that computes a hidden activation
$h_t$ combining it with the hidden activation at the previous step $h_{t-1}$, 
and (iii) it projects
the hidden activation onto a space of dimensionality equal to the vocabulary size $V$, 
followed by a \emph{softmax} function that turns the output vector
into a valid probability distribution. More formally, given a binary column vector $x_t$ representing the input symbol at step
$t$, we retrieve its corresponding embedding $w_t$ through $w_t = W_mx_t$, where $W_m$ is the embedding matrix with dimensionality $\mathbb{R}^{M x V}$. The hidden state in the standard RNN is given by
\begin{equation}
h_t = \sigma(W_{ih} w_i + W_{hh} h_{t-1} + b_h)
\end{equation}
\noindent where $W_{ih}$ and $W_{hh}$ are respectively the input-to-hidden and hidden-to-hidden 
projection matrices with dimensionality $\mathbb{R}^{M x H}$ and $\mathbb{R}^{H x H}$,
$b_h$ is a bias vector and $\sigma$ is the sigmoid non-linear function.
Finally, the probability assigned to each entry in the vocabulary at step $t$ is defined by the \emph{softmax}
\begin{equation}
P_{t,j} = \frac{e^{o_{t,j}}}{\sum_k^V e^{o_{t,k}}}
\end{equation}
where $o_{t,j}$ is the jth entry in the output vector $o_t = W_{ho}h_t$
and $W_{ho}$ is the hidden-to-output projection with dimensionality $\mathbb{R}^{HxV}$.

In practice, training an RNN is difficult due to the \emph{vanishing gradient problem} \cite{Hochreiter1998}
that makes it hard to apply the back-propagation algorithm for parameter tuning over
long sequences. Therefore, it is common to implement the recurrent layer using an enhanced
RNN like, e.g. Long Short-term Memory (LSTM) \cite{Hochreiter1997}. An LSTM-based RNNLM only differs from the 
previous RNNLM in the way the hidden activation $h_t$ is computed.
An LSTM cell incorporates three learnable gates---an input, forget and output gate---of shape:
\begin{equation}
i_t = \sigma(W_{ih}^iw_t + W_{hh}^i + b_h^i)
\label{eq:lstm_i}
\end{equation}
\begin{equation}
f_t = \sigma(W_{ih}^fw_t + W_{hh}^f + b_h^f)
\label{eq:lstm_f}
\end{equation}
\begin{equation}
o_t = \sigma(W_{ih}^ow_t + W_{hh}^o + b_h^o)
\label{eq:lstm_o}
\end{equation}
\noindent where $W^i$, $W^f$ and $W^o$ are, respectively, the gates parameters, 
and a writable memory cell $c_t$ that is updated following
\begin{equation}
c_t = f_t \odot c_{t-1} + i_t \odot \tanh(W_{ih}^c w_t + W_{hh}^c h_{t-1} + b_h^c)
\label{eq:lstm_c}
\end{equation}
(where $\odot$ is element-wise product and $\tanh$ is the hyperbolic tangent non-linear function). Finally, the memory cell is combined with the output gate to yield the hidden activation $h_t$: $h_t = o_t \odot \sigma(c_t)$\label{eq:lstm_h_t}. As can be seen from the equations, the role of the gates is to learn to write to and delete from the memory cell based on the input (Equations \ref{eq:lstm_f}, \ref{eq:lstm_i} and \ref{eq:lstm_c}), as well as to use the memory cell to compute the hidden activation (Equation \ref{eq:lstm_h_t}).

A RNNLM has as parameter the embedding matrix $W^m$, the hidden-to-output projection
$W_{ho}$, as well as the input-to-hidden and hidden-to-hidden projections of the RNN/LSTM networks. Theoretically, what sets a RNNLM apart is that it consistently displays a much larger context awareness---because of its ability to carry over information in the hidden state across very large spans---and that it is therefore able to learn syntactic dependencies and structures from the training material.
This is in stark contrast with a NGLM, which only reasons on the basis of a very local history and have little abstractive power.

Importantly, however, it should be emphasized that most approaches to AA operate on very \emph{local features}, such as lower-order character ngrams \cite{stamatatos_law,sapkota-EtAl:2015,kestemont:2014}. Most state-of-the-art models for AA indeed depend on document vectors containing normalized character ngram frequencies (typically in the range of 2-4), which are fed to a standard classifier, such as a support-vector machine with a linear kernel. The fact that the RNNLM might generate more realistic sentences than the NGLM does not necessarily entail that it would have an advantage in AA with respect to a conventional NGLM, which will stay closer to the original source documents. An important, if only secondary, question is therefore whether the use of an RNNLM in the context of AA would outperform a conventional NGLM, even if only very local features, such as character ngrams, are included in the model.

\section{Experimental setup}\label{sec:exp}
\subsection{Design}
The \textit{Patrologia Latina} (PL) is a corpus containing texts of Latin ecclesiastical writers in 221 volumes ranging a time span of 10 centuries, from Late Antiquity to  the High Middle Ages (3rd-13th century). It was first published in two series halfway the 19th century by Jacques-Paul Migne, who mainly based the texts off of 17th and 18th-century prints. Its digitized version is available since 1993, and it has remained one of the most sizable Latin corpora online ($\pm$113M words). 

Performing this experiment on the PL, and not on an English corpus, for instance, has been a conscious decision to raise the bar. It has been observed that state-of-the-art AA on an inflected language such as Latin yields poorer results when it is reliant on most frequent words \cite{rybicki2011}. Moreover, the Latin that has come down to us from the 1st century AD onwards is an institutionalized literary language, hardly a natural language, showing only far resemblance, or occasionally no resemblance at all, to the writer's mother tongue \cite{maes2009}. Tracing stylistic properties within a heavily formalized language, and attempting to resuscitate these through generation, is therefore challenging. An additional obstacle for both language generation as AA is that many of the PL's authors cite from similar, authoritative sources such as the Bible or the church fathers' precursory texts, thereby having in common an ecclesiastical vocabulary that could complicate the detection of stable writing style patterns.

\begin{figure}[h!]
    \hspace*{-0.5cm}
    \includegraphics[width=.5\textwidth]{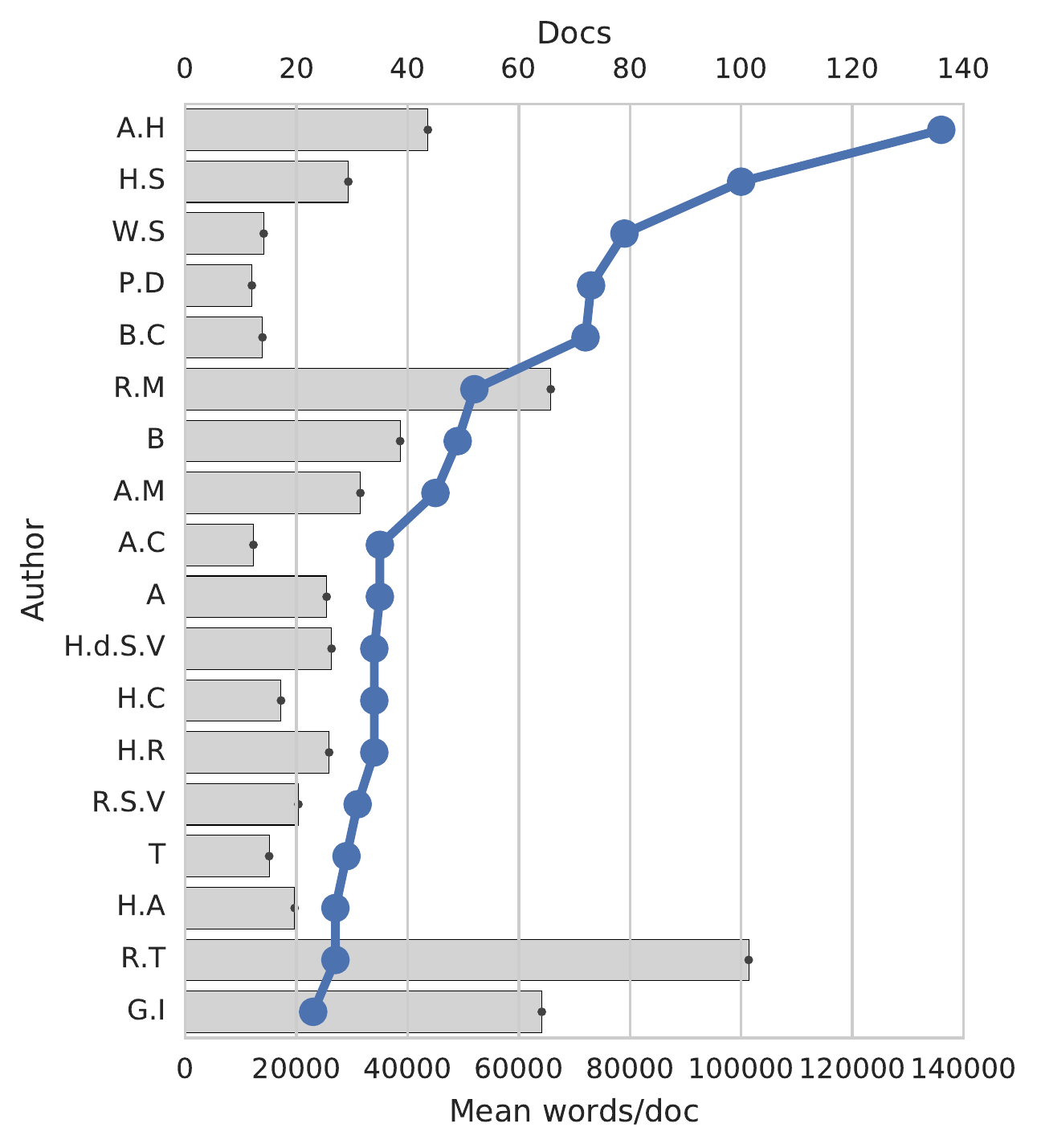}
    \caption{
	18 most prolific \textit{Patrologia Latina} authors ranked by document count. The bars yield an average of the document length.
    }
    \label{fig:pl}
\end{figure}

Not all authors in the PL have been equally prolific. These circumstances considerably limit the set of authors for whom our task is suited \cite{eder2015}. We set the condition that our text data include only texts by authors who dispose of at least 20 authentic, individual documents each. As such we favored document counts over token counts, and lexical variety over mere word quantity. A list of the 18 most prolific authors, their number of documents and the respective average length of these documents is given in Fig. \ref{fig:pl}.

\begin{figure}[h]
	\centering
	\includegraphics[width=.45\textwidth]{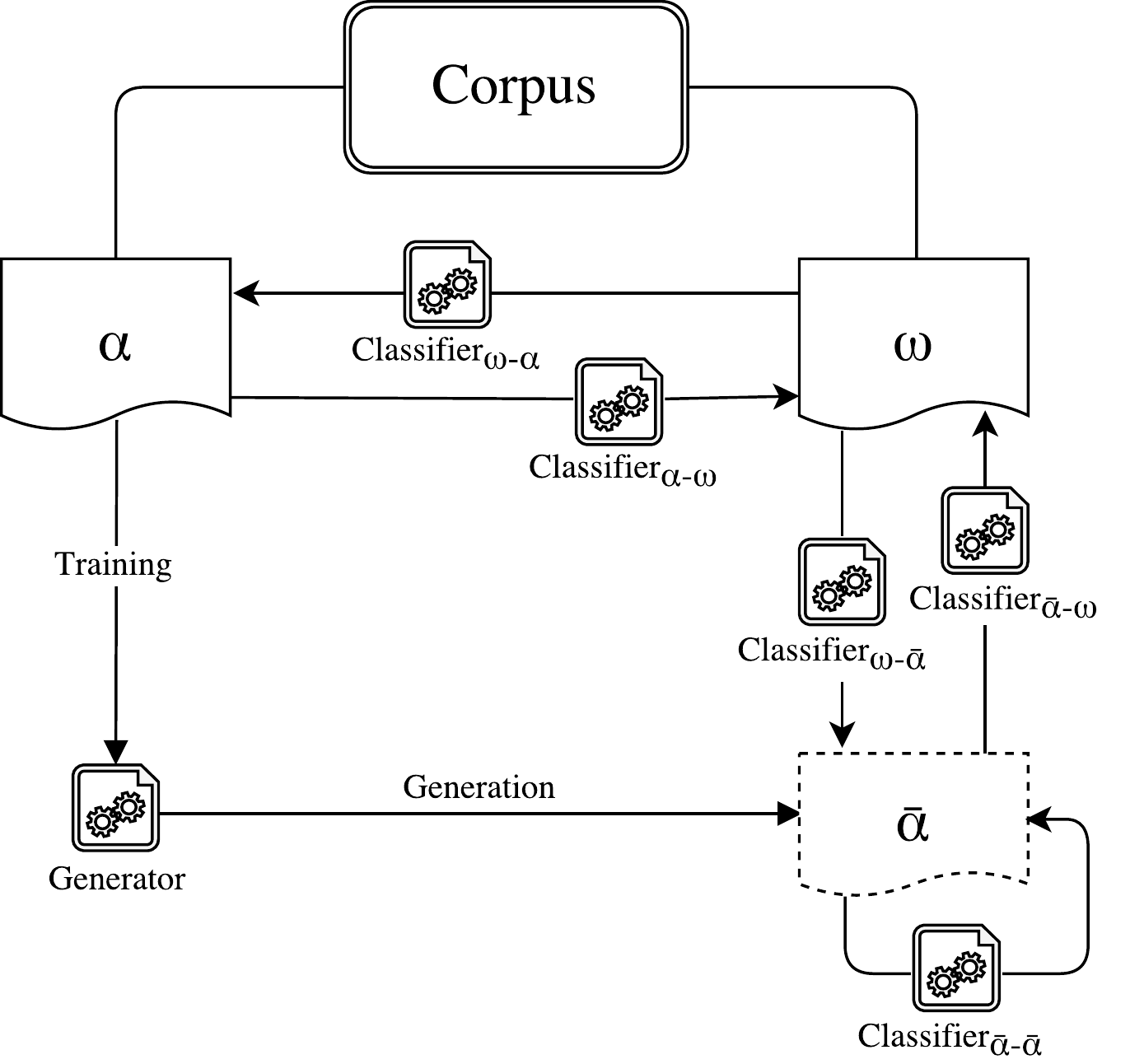}
    \caption{
    Experimental setup. $\protect\alpha$ and $\protect\omega$ refer
    to 50\% splits of the full corpus. $\protect\bar{\alpha}$
    refers to the generated dataset (cf. dashed
    line). Each classifier symbol refers to a classification
    experiment using the data at the arrow's source (first subscript)
    for training and the data at the arrow's target (second
    subscript) for testing (note that training only has to be
    performed 3 times, one per dataset).
    }
    \label{fig:setup}
\end{figure}

It is not trivial to design an experiment that allows us to study the behavior
of generated text in the context of AA. Fig. \ref{fig:setup} shows the experimental setup which we propose, and in which we attempt to maximize the
comparability of both generated and authentic data. We start by splitting the full
corpus into two equal-size document collections (stratified at the author level), $\alpha$ and $\omega$.
Only $\alpha$ will be used to train a LM, which then generates a third collection of synthetic documents.
For each author in $\alpha$
and $\omega$, we aggregate all documents into a list of sentences per sub-corpus. From these collections, we create 20 documents containing at least 5,000 words to create $\alpha$ and $\omega$, through randomly sampling sentences (without replacement) from the author's sentence collection. For the creation of $\bar{\alpha}$, we would also create 20 artificial 5,000-word documents, but this time through sampling new sentences from the LM. This approach has its limitations, because we limit and balance the available data to a considerable extent. Furthermore, the sampling procedure implies an underestimation in attribution performance, since it strips away all supra-sentential information.
Nevertheless, this setup guarantees that the authentic and 
generated corpora are maximally comparable in terms of number of documents, document length, topical diversity and style mixture---which is our focus in the present study.

Subsequently, 5 classification experiments are defined, where we train and and test on different 2-way 
combinations of the 3 datasets. In a first pair of experiments, $<\alpha,\omega>$ and $<\omega,\alpha>$, we
train and test a classifier on the authentic datasets to assess the classifier's performance under natural
conditions. (Note that we apply the classifier in both directions to account for any directionality artifacts.)
In a third experiment, we train and test a classifier on the generated data only ($<\bar{\alpha},\bar{\alpha}>$) to establish
to which extent the generated data preserves the data's stylistic structure at the author level 
(i.e. auto-classification). Fourthly, we conduct an experiment where we train on the generated data 
in $\bar{\alpha}$ and test on the authentic data in $\omega$ ($<\bar{\alpha},{\omega}>$). This allows 
us to verify whether the generated documents retain enough stylistic information to correctly attribute
authentic documents. Finally, we train a classifier on the authentic data in $\omega$ and test it on $\bar{\alpha}$: this setup ($<{\omega},\bar{\alpha}>$) allows to assess whether a classifier, trained on authentic data is still able to correctly attribute the generated materials. 

In addition, we conduct a final experiment which can be characterized from the point of view of self-learning or co-learning \cite{mihalcea2004co}---a semi-supervised learning technique where a core of training data is expanded with examples from a related but unlabeled dataset that can be classified with high confidence by a classifier trained in the original labeled dataset. In this experiment we compare the NGLM and RNNLM models with respect to their capacity to boost attribution performance by adding synthetic examples to the original training set---which might be a valuable strategy for real-life experiments. Specifically, we perform attribution on $\omega$ using a combination of $\alpha$ and $\bar{\alpha}$ as training data ($<\alpha+\bar{\alpha},\omega>$).


\subsection{Language Model Architectures for Text Generation}
In Section \ref{sec:cltg}, the text-generation and model parameters were defined. 
For the present experiments we generate 20 documents of 5000 words each using
a $\tau$ value of 1 and a $m$ value estimated on each author's dataset. For the RNNLM we reset the seed (parameter $s$) every 10 successfully generated
sentences, whereas for the NGLM we do it after every sentence. This asymmetry is motivated by the fact that NGLM the output distribution of an NGLM at each step is much more skewed and therefore sentences generated from the same seed tend to be be much less varied. For model fitting we set the NGLM order at 6, which, on a subjective evaluation, seemed a sufficiently large value for the comparatively small size of the datasets. 

For the RNNLM models the following parameter settings were selected. Embedding dimensionality $M$ was set to 24, the hidden layer dimension was 200 and we stacked up 2 LSTM layers to encourage the model to learn more abstract representations. Parameters were chosen based on common practice and reasonable defaults without further hyperparameter search. Each model was trained during 50 epochs using the adaptive variant of Stochastic Gradient Descent
Adam \cite{Kingma2015} with an initial learning rate of 0.001. We set a small batch size of 50, preferring stability over speed during training. Moreover, we clip the gradients before each batch update to a maximum norm value of 5 to avoid the exploding gradients following \cite{Pacanu2013} and truncate the gradient back-propagation after 50 recurrent steps. We also applied 30\% dropout after each recurrent layer following \cite{Zaremba2015} to avoid overfitting. For each RNNLM we held out a validation set using 10\% of the data to monitor and evaluate training.
We ensured that validation perplexity was always lower than train perplexity.
Average validation perplexity was 4.015 with a standard deviation of 0.183.\footnote{
	All software associated with this paper is available from \url{https://www.github.com/jedgusse/project_lorenzo}.
}

\subsection{Attribution as Classification}\label{sec:txtsvm}
For the AA classification as described in the experimental setup of section \ref{sec:exp}, we use a linear SVM classifier \cite{diederich2003}. 
We extract shallow linguistic features in the form of Tfidf-weighted character ngrams (from bigrams to four-grams) as style markers by which to determine authorship. Note that the feature extraction of ngrams in the order of 2 to 4 might have important repercussions, since NGLM training fully focuses on capturing that particular distribution, whereas the more expressive RNNLM models full sequences. Furthermore, we refrain from using word-level features such as word ngrams or POS tags, since this would introduce a further asymmetry in the comparison given that the RNNLM can generate unseen words whereas the NGLM can not.
The model accuracy of the SVM is finetuned by searching over different value ranges for the SVM's parameters. 
The number of features is set to range from 1,000 to 30,000 max features for each fit, more specifically in the following order: 5,000, 10,000, 15,000 and 30,000 features. For the C-parameter of the SVM we search over values of respectively 1, 10, 100 and 1,000.

\section{Results}

\begin{figure*}[t]
	\centering
	\includegraphics[width=\textwidth]{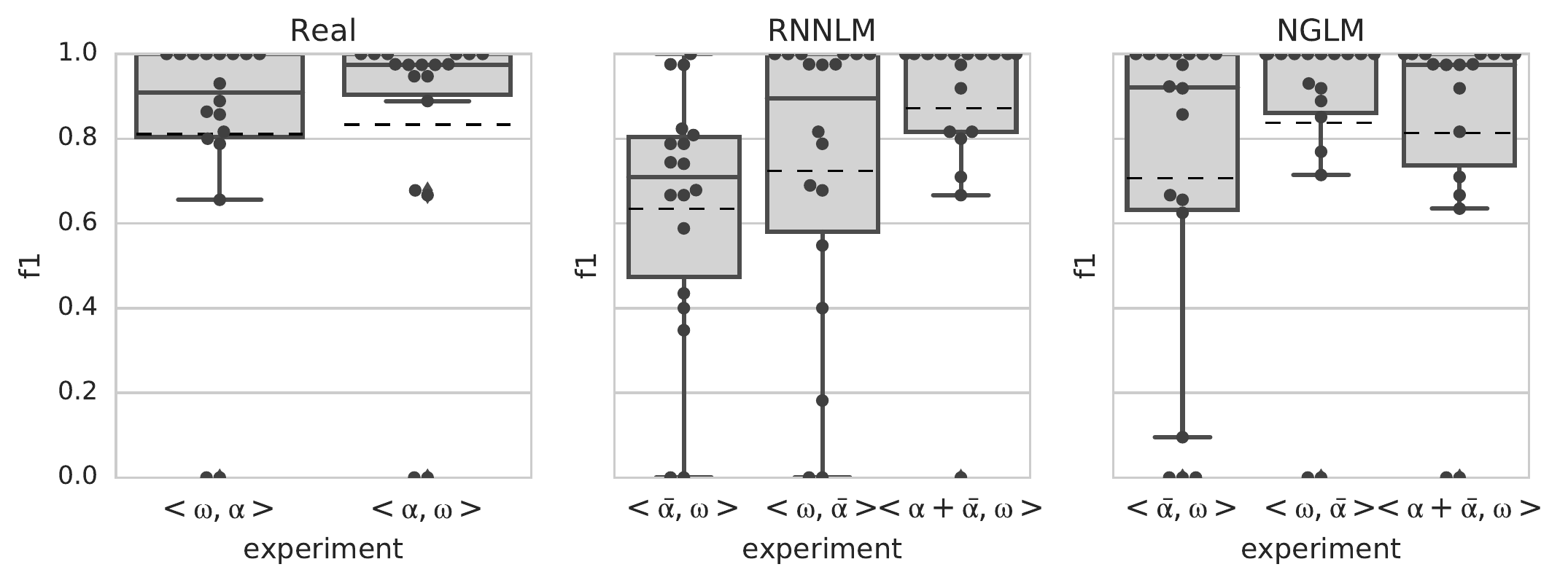}
   	\caption{
	F1 scores for the different combinations of $\alpha$, $\bar{\alpha}$, and $\omega$.
   	}
   	\label{fig:f1}
\end{figure*}

\begin{table}
	\hskip-0.5cm\begin{tabular}{llccc}
	\toprule
	Source & Experiment & F1 & P & R \\
	\midrule
	Real & $<\alpha,\omega>$ &  0.833 &      0.818 &   0.869 \\
   	  	 & $<\omega,\alpha>$ &  0.811 &      0.795 &   0.853 \\
	NGLM & $<\alpha+\bar{\alpha},\omega>$ &  0.814 &      0.809 &   0.850 \\
    	  & $<\bar{\alpha},\omega>$ &  0.706 &      0.744 &   0.750 \\
     	  & $<\omega,\bar{\alpha}>$ &  0.837 &      0.811 &   0.881 \\
	RNNLM & $<\alpha+\bar{\alpha},\omega>$ &  0.872 &      0.878 &   0.892 \\
        & $<\bar{\alpha},\omega>$ &  0.635 &      0.701 &   0.658 \\
        & $<\omega,\bar{\alpha}>$ &  0.724 &      0.778 &   0.775 \\
	\bottomrule
   
	\end{tabular}
\caption{\label{tab:results}
	Mean F1, Precision (\emph{P}) and Recall (\emph{R}) scores for all classification experiments.
}
\end{table}

\subsection{Examples of Generated Language}
What follows are two short extracts from the respective outputs of an NGLM and RNNLM trained on Augustine (A.H.) (most prolific author of the dataset, see Table \ref{fig:pl}), which gives an anecdotal intuition of how the output of these language models differs.\\

\noindent \textbf{Ngram-based LM $(\bar{\alpha}$)}
\begin{exe}
	\ex[*]{
	\gll Sed uis uenire: \textbf{quod} postridie, ascensiones honora pastorem, nec sane reipublicos idem testis et implebitur tamen mentiendum sit propitiaberis. \\
		Yet {you wish} {to come} since tomorrow ascensions honoured  {the shepherd}, {and not} completely republican {the same} witness also {will be fulfilled} nevertheless {to be deceived} {it may be} {you will be enriched}.\\
	\trans{}}
\end{exe}

\noindent \textbf{RNN-based LM} $(\bar{\alpha}$)
\begin{exe}
	\ex[*]{
	\gll Et idam precepti, siue ad sensum noui: nonuulde sunt enim Filius Domini substantia, sed non sunt \textbf{qui} secururum superbia et \textbf{perrectus} est, mortalis includendi estus que fiumus propter illam uideantur. \\
    And {that same (?)} commandment, {be it} towards {the feeling} {I know}: {not enough (?)} are {after all} {the Son} {of our Lord} {our substance}, but none {are there} who {amongst the untroubled} {through pride} also righteous are, mortal {by including} fire and {we were (?)} {because of} this {may they be beheld}.\\
	\trans{}}
\end{exe}

\noindent The extract of RNNLM-generated text as compared to the NGLM demonstrates how the RNNLM is better at reproducing a syntactic logic (which moreover makes translation easier). Note, for instance, how the nominative of the relative pronoun \textit{qui} is maintained towards the end of the subordinate clause in the participle perfect \textit{perrectus}, and even seems to be carried on in the next clause as opposed to the awkwardly placed \textit{quod} in the ngram-based extract. The RNNLM is also arguably better at positioning the verbs in the clauses. Compare, for instance, the NGLM's dense verbal sequence \textit{implebitur tamen mentiendum sit propitiaberis}. Finally, the RNNLM is more apt at generating plausible neologisms. Examples include \textit{idam} (cfr. \textit{idem} and \textit{quidam}), \textit{fiumus} (cfr. \textit{fiemus}), secururum (cfr. \textit{securus} and the genitive ending \textit{-orum} and \textit{-arum}). To a human reader, the RNNLM produces superficially more convincing text.

\subsection{Attribution results}
The results of the attribution experiments are presented in Table \ref{tab:results} in terms of recall, precision and F1-scores and the distributions are visualized in Fig. \ref{fig:f1}. We focus on the macro-averaged F1-scores in our discussion, although one should not forget that the scores vary considerably over individual authors (cf. Fig. ~\ref{fig:f1}). With respect to the authentic data, classifying $\alpha$ on the basis of $\omega$ is slightly more difficult than the reverse direction, which seems a negligible directionality artifact. When we use the generated data as training material to classify authentic material $<\bar{\alpha},\omega>$ , we see that the F1-scores drop significantly for both LMs, although the NGLM seems more robust in this respect. Interestingly, the drop is much less significant for the opposite situation, where we train on authentic material and classify generated material $<\omega, \bar{\alpha}>$. This suggests that enough stylistic information is preserved in the generated text to attribute it to the original author, but that this information in isolation does not suffice to train a convincing attribution system on. When used in isolation, the NGLM outperforms the RNNLM in both setups. However, the situation is clearly different for the augmentation or self-learning setup ($<\alpha+\bar{\alpha},\omega>$)---c.f. Section \ref{sec:exp}---, where we train an attributor on the combination of $\alpha$ and $\bar{\alpha}$, and test it on the authentic $\omega$ set. Here, we see that the RNNLM performs better than the NGLM in the corresponding experiment -- the NGLM in fact even performs worse in this case than in the normal $<\alpha,\omega>$ setup.

\subsection{Discussion}
\begin{figure}[ht]
\centering
\begin{subfigure}{9cm}
	\centering\hspace*{-2cm}
	\includegraphics[width=\linewidth, scale=0.2]{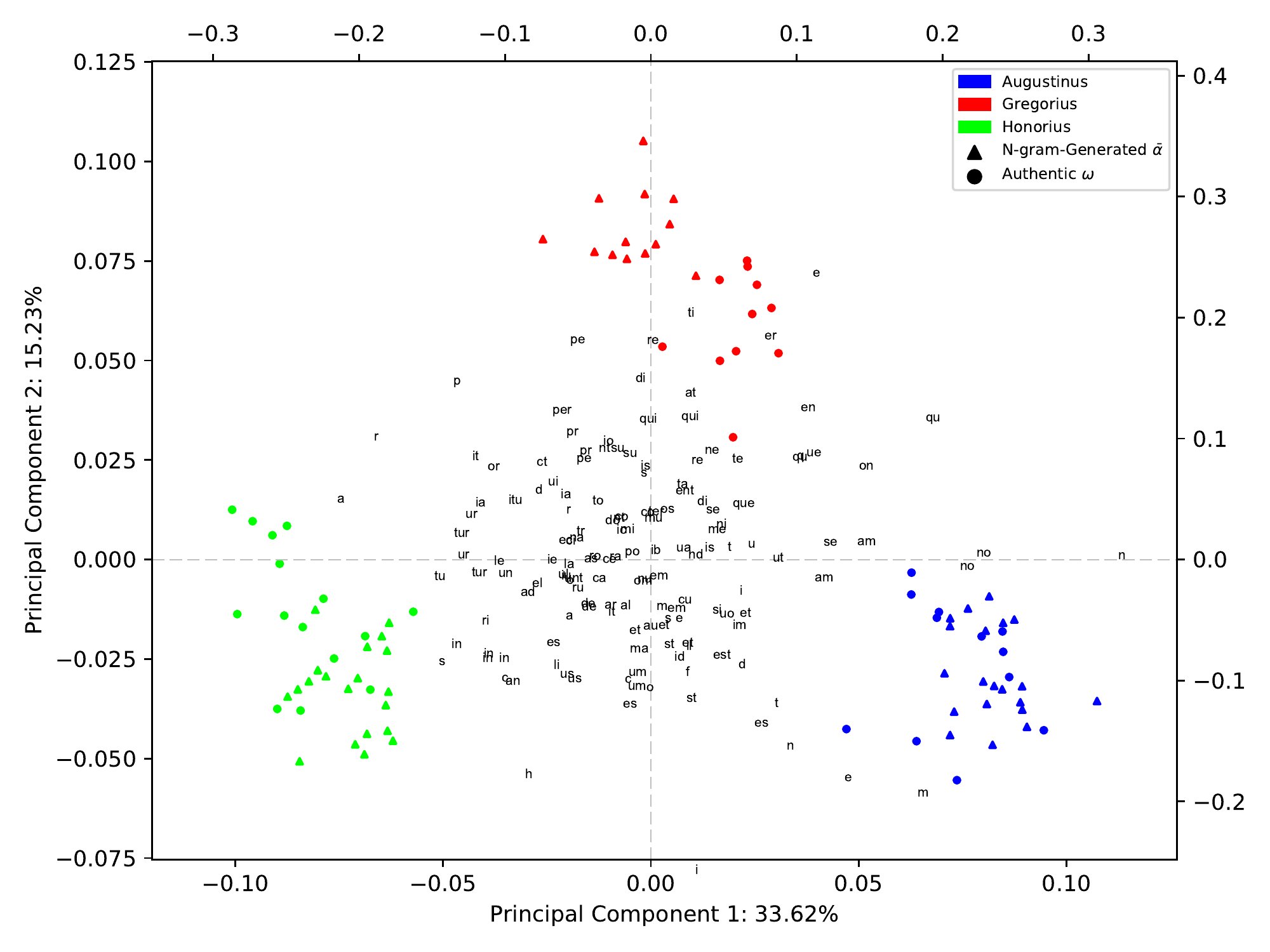}
\end{subfigure}

\begin{subfigure}{9cm}
	\centering\hspace*{-2cm}
	\includegraphics[width=\linewidth, scale=0.2]{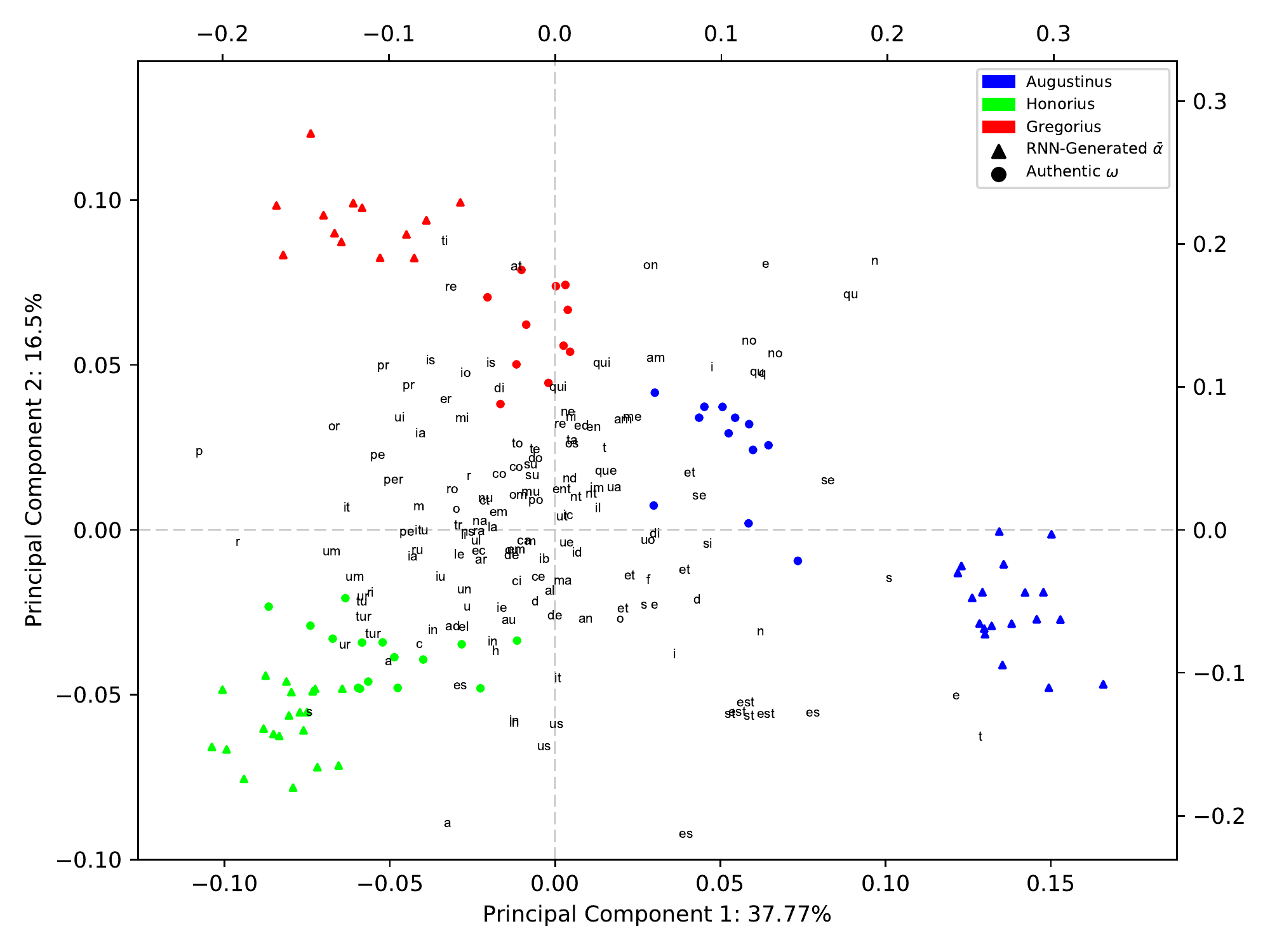}
    \end{subfigure}
    \caption{PCA plots (1st 2 PCs) for 3 authors using document vectors representing the normalized frequencies of the 150 most frequent ngrams (order 2-4) in 2500-word sample. We include a mixture of authentic $\omega$ data and generated $\bar{\alpha}$ data (\emph{top: NGLM; bottom: RNNLM}).}
    \label{fig:pca}
\end{figure}

To understand the difference in behavior between both LMs, it is useful to inspect Fig.~\ref{fig:pca}. Here, we use a Principal Components Analysis  \cite{binongo1999} to visualize 2500-word samples for 3 three most prolific authors (Augustine of Hippo, Honorius of Autun, and Gregory the Great) using the 150 most common ngrams. We include a mixture of authentic $\omega$ data and generated $\bar{\alpha}$ data for each author, comparing the NGLM and the RNNLM. The plots shows that NGLM produces text samples which lie very close in ngram frequencies to the authentic data, whereas the texts produced by the RNNLM follow a markedly different distribution than $\omega$ -- this difference is very outspoken for Augustine, for instance. As might be expected on the basis of the observation in section \ref{sec:txtsvm}, the NGLM thus produces data that stays very close to the original input, whereas RNNLM yields fuzzier texts, that follow a slightly different distribution. This explains why it is, for example, easier to train an attributor on the data generated by an NGLM than an RNNLM.

Conversely, our results show that the situation is different in the data augmentation setup, where we train an attributor on the combination of $\alpha$ and $\bar{\alpha}$ and test it on the authentic $\omega$ set. In this case, the NGLM performs worse than in the corresponding the non-augmented setup, whereas the performance of the RNNLM sensitively increases. Arguably, the fuzziness of the RNNLM-generated data adds an interesting complexity to the original core of authentic data, which can be exploited by the classifier.

\begin{figure}[h!]
    \hspace*{-0.5cm}
    \includegraphics[width=.5\textwidth]{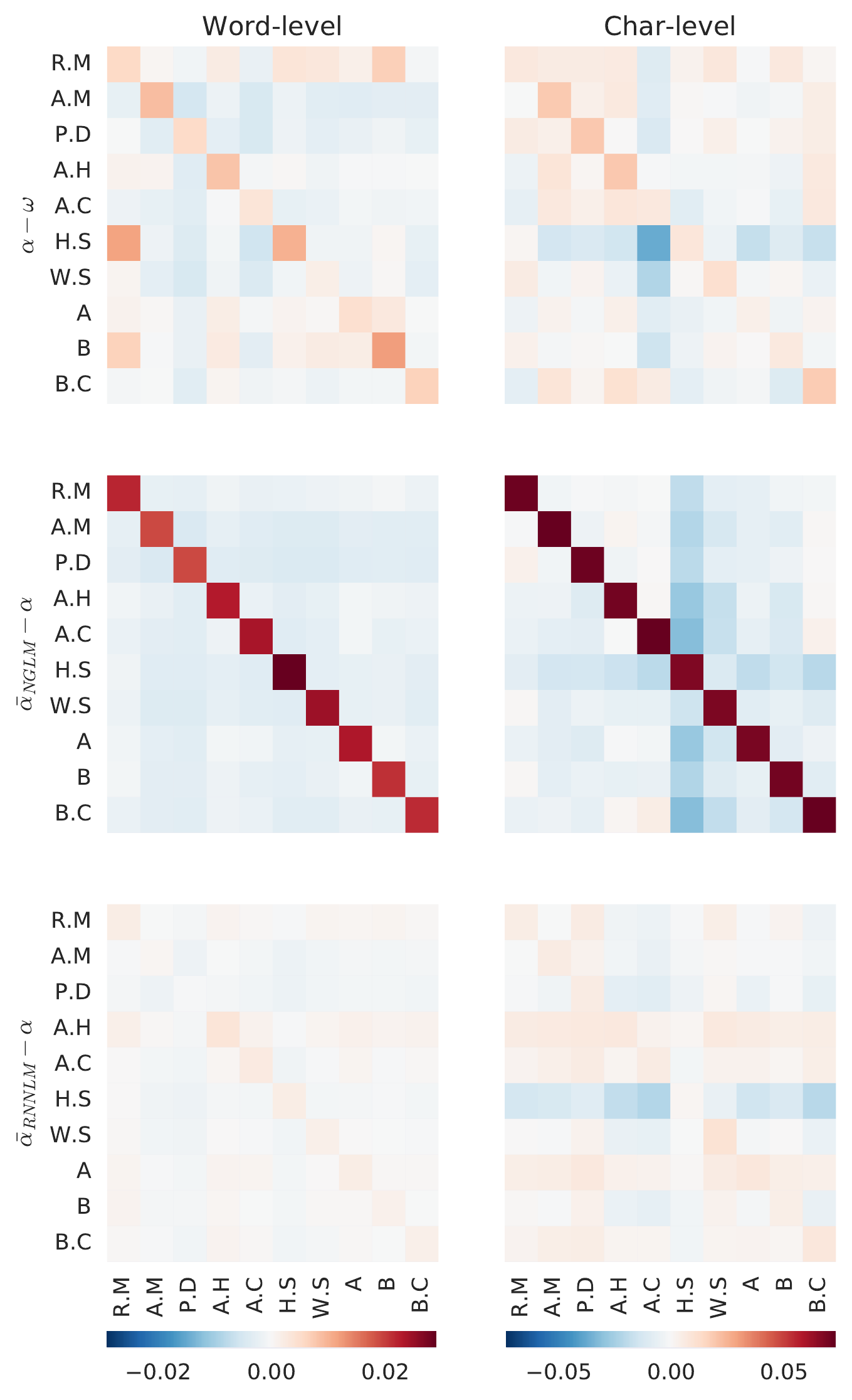}
    \caption{
	Mean-normalized Jaccard similarity scores between 10 most prolific authors using word (left column) and character (right column) bigrams to fourgrams, comparing real data (first row), RNNLM-synthetic data with real data (second row) and NGLM-synthetic data with real data (third row).
    }
    \label{fig:jaccard}
\end{figure}

While these results indirectly show that the RNNLM did not simply overfit on $\alpha$, it is an interesting question to which extent $\alpha$ and $\bar{\alpha}$ display (lexical) overlap in the case of both LMs. If the overlap would indeed be larger for the NGLM than the RNNLM, this would support our interpretation. In Fig.~\ref{fig:jaccard}, we show mean-normalized, pairwise Jaccard similarities for the 10 most prolific authors in both $\alpha$ and $\bar{\alpha}$ for each LM. The dark diagonals in the second row of the heatmaps visually support the observation that the NGLM displays a much more outspoken overlap between $\alpha$ and $\bar{\alpha}$. Such an effect is much more faint in the case of the RNNLM and in this respect it remains more faithful to the real data (first row for $\alpha$ and $\omega$).

\section{Conclusion}
Our preliminary results confirm that the texts generated by a traditional NGLM are
relatively `dull' and `conservative' in the sense that they stay relatively close
to the local distribution of the source data on which they were trained. Conceptually, the RNNLM has a clear advantage in terms of expressiveness and capacity with respect to the NGLM. In practice, given the small size of AA datasets, an underfitted RNNLM yield fuzzier examples, which explains why the NGLM outperforms the RNNLM when the classifier is \emph{restricted} to the generated data ($<\bar{\alpha},\omega>$ and $<\omega,\bar{\alpha}>$). At the same time, the training data augmentation setup ($<\alpha + \bar{\alpha},\omega>$) shows that whereas NGLM-generated data adds comparatively little to the authentic data---reproducing a subset of the original feature distribution, as shown in Fig. \ref{fig:jaccard}---, the RNNLM-generated data presents a valuable data contribution which does result in an absolute increase in attribution performance with respect to the real classification setup $<\alpha, \omega>$. Although further research into the matter is needed, this clearly suggests that the complexity of the RNNLM data is useful for training data augmentation, arguably capturing stylistic nuances which a simpler LM cannot.

In the future, we will explore the flexibility of the general RNNLM  framework to develop generative architectures that better capture the style of the training data. In particular, following \cite{Linzen2016} we hypothesize that forcing the RNN to model more linguistic structure---e.g. jointly modeling words and POS-tags---, should result in better language generation and better style preservation. Furthermore, we plan on exhaustively testing the capabilities of author-specific generative models for self-learning in AA, investigating the effect of adding different amounts of synthetic data and selectively adding synthetic data based on the confidence with which it can be correctly classified by a classifier trained on real data.

Additionally, we would like to investigate pre-training in out-of-domain data as well as more compact ways of modelling author-specific language---such as conditional language models \cite{Tang2016}---as means to alleviate underfitting of the RNN models on small datasets.

\bibliographystyle{acl}{}
\bibliography{acl2015}

\newpage

\appendix
\section{Author names with abbreviations}

\begin{tabular}{ll}
\toprule
    Abbrv &                     Author \\
\midrule
      H.S &    Hieronymus Stridonensis \\
      G.I &                Gregorius I \\
      A.H &     Augustinus Hipponensis \\
      A.M &    Ambrosius Mediolanensis \\
      B &                       Beda \\
      H.C &   Hildebertus Cenomanensis \\
      H.d.S.V &         Hugo de S- Victore \\
      R.T &        Rupertus Tuitiensis \\
      W.S &          Walafridus Strabo \\
      T &               Tertullianus \\
      P.D &            Petrus Damianus \\
      H.A &   Honorius Augustodunensis \\
      H.R &        Hincmarus Rhemensis \\
      B.C &  Bernardus Claraevallensis \\
      A &                   Alcuinus \\
      R.M &             Rabanus Maurus \\
      A.C &     Anselmus Cantuariensis \\
      R.S.V &      Richardus S- Victoris \\
\bottomrule
\end{tabular}

\end{document}